\documentclass[runningheads]{llncs}

\usepackage{color}
\usepackage{amsmath}
\usepackage{amssymb}
\usepackage{mathtools}
\usepackage{stmaryrd}
\usepackage[ruled,vlined,linesnumbered]{algorithm2e}
\usepackage{url}
\usepackage{multirow}
\usepackage{float}
\usepackage{caption}
\usepackage{subcaption}
\newcommand{\comment}[1]{$\triangleright$ #1}   % For usage inside pseudo code
\newcommand{\sect}{Section}
\newcommand{\alg}{Alg.}
\newcommand{\algln}{line}
\newcommand{\alglns}{lines}
\newcommand{\tab}{Table}
\newcommand{\fig}{Fig.}

\newcommand{\labl}{\lambda}             % A single label
\newcommand{\iterlabels}{k}             % The variable that iterates the labels
\newcommand{\numlabels}{K}              % The total number of available labels
\newcommand{\labelset}{\mathcal{L}}     % The set of available labels
\newcommand{\labelspace}{\mathcal{Y}}   % The label space, i.e., the set of all possible labelings

\newcommand{\truelabel}{y}              % A single label according to the ground truth
\newcommand{\truelabelvect}{\vec{y}}    % A label vector according to the ground truth
\newcommand{\truelabelmatr}{Y}          % A matrix of labels according to the ground truth

\newcommand{\predlabel}{\hat{y}}        % A predicted label
\newcommand{\predvect}{\vec{\hat{y}}}   % A predicted label vector
\newcommand{\predmatr}{\hat{Y}}         % A matrix of predicted labels

\newcommand{\conf}{\hat{p}}             % A predicted score
\newcommand{\confvect}{\vec{\hat{p}}}   % A vector of predicted scores

\newcommand{\ex}{\vec{x}}               % A single example
\newcommand{\iterex}{n}                 % The variable that iterates the examples
\newcommand{\numex}{N}                  % The total number of available examples

\newcommand{\attr}{A}                   % A single numerical or nominal attribute
\newcommand{\iterattr}{l}               % The variable that iterates the attributes
\newcommand{\numattr}{L}                % The total number of available attributes
\newcommand{\attrval}{x}                % The value that is associated with an attribute
\newcommand{\attrspace}{\mathcal{X}}    % The attribute/example space

\newcommand{\dataset}{\mathcal{D}}      % A (training) data set

\newcommand{\classifier}{f}                 % A single classifier
\newcommand{\ensemble}{F}                   % An ensemble, i.e., a set of classifiers
\newcommand{\iterclassifiers}{t}            % The variable that iterates the members of an ensemble
\newcommand{\numclassifiers}{T}             % The total number of classifiers in an ensemble
\newcommand{\classifierspace}{\mathcal{F}}  % The space of potential classifiers

\newcommand{\rul}{\classifier}          % A single rule
\newcommand{\body}{b}                   % The body of a rule
\newcommand{\head}{\confvect}           % The head of a rule
\newcommand{\cond}{c}                   % A single condition
\newcommand{\shrinkage}{\eta}           % The shrinkage parameter

\newcommand{\measure}{\mathcal{L}}      % An evaluation measure
\newcommand{\loss}{\ell}                % A loss function

\newcommand{\objective}{\mathcal{R}}                    % The training objective
\newcommand{\approxobjective}{\widetilde{\mathcal{R}}}  % The approximate training objective
\newcommand{\regterm}{\Omega}                           % A regularization term
\newcommand{\regweight}{\lambda}                        % The L2 regularization weight

\newcommand{\gradient}{g}                               % A single gradient
\newcommand{\gradientvect}{\vec{g}}                     % A vector of gradients
\newcommand{\gradientset}{\mathcal{G}}                  % A set of gradient vectors
\newcommand{\hessian}{h}                                % A single Hessian
\newcommand{\hessianmatr}{H}                            % A matrix of Hessians
\newcommand{\hessianset}{\mathcal{H}}                   % A set of Hessian matrices

\newcommand{\iterrows}{i}   % The variable that iterates the rows of a matrix
\newcommand{\itercols}{j}   % The variable that iterates the columns of a matrix

\DeclareMathOperator{\sgn}{sgn}             % The sign function
\DeclareMathOperator{\logistic}{logistic}   % The logistic function
\DeclareMathOperator{\diag}{diag}           % Build diagonal matrices from vectors
\DeclareMathOperator*{\argmin}{arg\,min}    % The argmin operator

\begin{document}

\title{Learning Gradient Boosted Multi-label Classification Rules}
\titlerunning{Learning Gradient Boosted Multi-label Classification Rules}

\author{Michael Rapp\inst{1} \and Eneldo Loza Menc\'ia\inst{1} \and Johannes F\"urnkranz\inst{2} \and\\ Vu-Linh Nguyen\inst{3} \and Eyke H\"ullermeier\inst{3}}
\authorrunning{M. Rapp et al.}

\institute{Knowledge Engineering Group, TU Darmstadt, Germany\\
\email{mrapp@ke.tu-darmstadt.de}, \email{research@eneldo.net}
\and
Computational Data Analysis Group, JKU Linz, Austria\\
\email{juffi@faw.jku.at}
\and
Heinz Nixdorf Institute, Paderborn University, Germany\\
\email{vu.linh.nguyen@uni-paderborn.de}, \email{eyke@upb.de}
}

\maketitle

\begin{abstract}
In multi-label classification, where the evaluation of predictions is less straightforward than in single-label classification, various meaningful, though different, loss functions have been proposed. Ideally, the learning algorithm should be customizable towards a specific choice of the performance measure. Modern implementations of boosting, most prominently gradient boosted decision trees, appear to be appealing from this point of view. However, they are mostly limited to single-label classification, and hence not amenable to multi-label losses unless these are label-wise decomposable. In this work, we develop a generalization of the gradient boosting framework to multi-output problems and propose an algorithm for learning multi-label classification rules that is able to minimize decomposable as well as non-decomposable loss functions. Using the well-known Hamming loss and subset 0/1 loss as representatives, we analyze the abilities and limitations of our approach on synthetic data and evaluate its predictive performance on multi-label benchmarks.

\end{abstract}

\section{Introduction}
\label{sec:intro}

Multi-label classification (MLC) is concerned with the per-instance prediction of a subset of relevant labels out of a predefined set of available labels. Examples of MLC include real-world applications like the assignment of keywords to documents, the identification of objects in images, and many more (see, e.g., \cite{tsoumakas2010} or \cite{zhang2013} for an overview). To evaluate the predictive performance of multi-label classifiers, one needs to compare the predicted label set to a ground-truth set of labels. As this can be done in many different ways, a large variety of loss functions have been proposed in the literature, many of which are commonly used to compare MLC in experimental studies. As these measures may conflict with each other, optimizing for one loss function often leads to deterioration with respect to another loss. Consequently, an algorithm is usually not able to dominate its competitors on all measures \cite{dembczynski2012a}.

For many MLC algorithms it is unclear what measure they optimize. On the other hand, there are also approaches that are specifically tailored to a certain loss function, such as the F1-measure \cite{pillai2017}, or the subset 0/1 loss \cite{nam2017}. \emph{(Label-wise) decomposable} losses are particularly easy to minimize, because the prediction for individual labels can be optimized independently of each other \cite{dembczynski2012a}. For example, binary relevance (BR) learning transforms an MLC problem to a set of independent binary classification problems, one for each label. On the other hand, much of the research in MLC focuses on incorporating label dependencies into the prediction models, which is in general required for optimizing \emph{non-decomposable} loss functions such as subset 0/1.

While BR is appropriate for optimizing the Hamming loss, another reduction technique, label powerset (LP), is a natural choice for optimizing the subset 0/1 loss \cite{dembczynski2012a}. The same applies to (probabilistic) classifier chains \cite{cheng2010,read2009}, which seek to model the joint distribution of labels. Among the approaches specifically designed for MLC problems, boosting algorithms that allow for minimizing multi-label losses are most relevant for this work. Most of these algorithms (e.g.\ \cite{zhang2019}, \cite{si2017}, \cite{johnson2005}, \cite{schapire2000} and variants thereof) require the loss function to be label-wise decomposable. This restriction also applies to  methods that aim at minimizing ranking losses (e.g.\ \cite{jung2017}, \cite{dembczynski2012b}, or \cite{schapire2000}) or transform the problem space to capture relations between labels (e.g. \cite{joly2019} or \cite{bhatia2015}). To our knowledge, AdaBoost.LC \cite{amit2007} is the only attempt at directly minimizing non-decomposable loss functions.

The first contribution of this work is a framework that allows for optimizing multivariate loss functions (\sect~\ref{sec:boosting}). It inherits the advantages of modern formulations of the gradient boosting framework, in particular the ability to flexibly use different loss functions and to incorporate regularization into the learning objective. The proposed framework allows the use of non-decomposable loss functions and enables the ensemble members to provide loss-minimizing predictions for several labels at the same time, hence taking label dependencies into account. This is in contrast to AdaBoost.LC, where the base classifiers may only predict for individual labels. Our experiments suggest that this ability is crucial to effectively minimize non-decomposable loss functions on real-world data sets.

Our second contribution is BOOMER, a concrete instantiation of this framework for learning multi-label classification rules (Section~\ref{sec:algorithm}). While there are several rule-based boosting approaches for conventional single-label classification, e.g., the ENDER framework \cite{dembczynski2010} which generalizes several rule-based boosting methods such as RuleFit \cite{friedman2008}, their use has not yet been systematically investigated for MLC. We believe that rules are a natural choice in our framework, because they define a more general concept class than the commonly used decision trees: While each tree can trivially be viewed as a set of rules, not every rule set may be encoded as a tree. Also, an ensemble of rules provides more flexibility in how the attribute space is covered. While in an ensemble of $\numclassifiers$ decision trees, each example is covered by exactly $\numclassifiers$ rules, an ensemble of rules can distribute the rules in a more flexible way, using more rules in regions where predictions are difficult and fewer rules in regions that are easy to predict.

\section{Preliminaries}
\label{sec:preliminaries}

In this section, we introduce the notation used throughout the remainder of this work and present the type of models we use. Furthermore, we present relevant loss functions and recapitulate to what extent their optimization may benefit from the exploitation of label dependencies.

In contrast to binary and multi-class classification, in multi-label classification an example can be associated with several class labels $\labl_{\iterlabels}$ out of a predefined and finite label set $\labelset = \left \{ \labl_{1}, \dots, \labl_{\numlabels} \right \}$. An example $\ex$ is represented in attribute-value form, i.e., it consists of a vector $\ex = \left( \attrval_{1}, \dots, \attrval_{\numattr} \right) \in \attrspace = \attr_{1} \times \dots \times \attr_{\numattr}$, where $\attrval_{\iterattr}$ specifies the value associated with a numeric or nominal attribute $\attr_{\iterattr}$. In addition, each example is associated with a binary label vector $\truelabelvect = \left( \truelabel_{1}, \dots, \truelabel_{\numlabels} \right) \in \labelspace$, where $\truelabel_{\iterlabels}$ indicates the absence ($-1$) or presence ($+1$) of label $\labl_{\iterlabels}$. We denote the set of possible labelings by $\labelspace = \left \{ -1, +1 \right \}^\numlabels$.

We deal with MLC as a supervised learning problem in which the task is to learn a predictive model $\classifier: \attrspace \to \labelspace$ from a given set of labeled training examples $\dataset = \left \{ \left( \ex_{1}, \truelabelvect_{1} \right), \dots, \left( \ex_{\numex}, \truelabelvect_{\numex} \right) \right \} \subset \attrspace \times \labelspace$. A model of this kind maps a given example to a predicted label vector $\classifier \left( \ex \right) = \left( \classifier_{1} \left( \ex \right), \dots, \classifier_{\numlabels} \left( \ex \right) \right) \in \labelspace$. It should generalize well beyond the given observations, i.e., it should yield predictions that minimize the expected risk with respect to a specific loss function. In the following, we denote the binary label vector that is predicted by a multi-label classifier as $\predvect = \left( \predlabel_{1}, \dots, \predlabel_{\numlabels} \right) \in \labelspace$.

\subsection{Ensembles of Additive Functions}
\label{sec:ensembles}

We are concerned with ensembles $\ensemble = \{ \classifier_{1}, \dots, \classifier_{\numclassifiers} \}$ that consist of $\numclassifiers$ additive classification functions $\classifier_{\iterclassifiers} \in \classifierspace$, referred to as \emph{ensemble members}. By $\classifierspace$ we denote the set of potential classification functions. In this work, we focus on classification rules (cf.~\sect~\ref{sec:rules}). Given an example $\ex_{\iterex}$, all of the ensemble members predict a vector of numerical confidence scores
\begin{equation}
\label{eq:confvect}
\confvect_{\iterex}^{\iterclassifiers} = \classifier_{\iterclassifiers} \left( \ex_{\iterex} \right) = \left( \conf_{\iterex 1}^{\iterclassifiers}, \dots, \conf_{\iterex \numlabels}^{\iterclassifiers} \right) \in \mathbb{R}^{\numlabels},
\end{equation}
where each score expresses a preference for predicting the label $\labl_{\iterlabels}$ as absent if $\conf_{\iterlabels} < 0$ or as present if $\conf_{\iterlabels} > 0$. The scores provided by the individual members of an ensemble can be aggregated into a single vector of confidence scores by calculating the vector sum
\begin{equation}
\label{eq:confvect_aggregated}
\confvect_{\iterex} = \ensemble \left( \ex_{\iterex} \right) = \confvect_{\iterex}^{1} + \dots + \confvect_{\iterex}^{\numclassifiers} \in \mathbb{R}^{\numlabels} \, ,
\end{equation}
which can subsequently be turned into the final prediction of the ensemble in the form of a binary label vector (cf.~\sect~\ref{sec:prediction}).

\subsection{Multi-label Classification Rules}
\label{sec:rules}

As ensemble members, we use conjunctive classification rules of the form
\[
\rul: \body \rightarrow \head \, ,
\]
where $\body$ is referred to as the \emph{body} of the rule and $\head$ is called the \emph{head}. The body $\body: \attrspace \to \left \{ 0, 1 \right \}$ consists of a conjunction of conditions, each being concerned with one of the attributes. It evaluates to $1$ if a given example satisfies all of the conditions, in which case it is said to be \emph{covered} by the rule, or to 0 if at least one condition is not met. An individual condition compares the value of the $\iterattr$-th attribute of an example to a constant by using a relational operator, such as $=$ and $\neq$ (if the attribute $\attr_{\iterattr}$ is nominal), or $\leq$ and $>$ (if $\attr_{\iterattr}$ is numerical).

In accordance with \eqref{eq:confvect}, the head of a rule $\head = \left( \conf_{1}, \dots, \conf_{\numlabels} \right) \in \mathbb{R}^{\numlabels}$ assigns a numerical score to each label. If a given example $\ex$ belongs to the axis-parallel region in the attribute space $\attrspace$ that is covered by the rule, i.e., if it satisfies all conditions in the rule's body, the vector $\head$ is predicted. If the example is not covered, a null vector is predicted. Thus, a rule can be considered as a mathematical function $\rul: \attrspace \to \mathbb{R}^{\numlabels}$ defined as
\begin{equation}
\rul \left( \ex \right) = \body \left( \ex \right) \head \, .
\end{equation}
This is similar to the notation used by Dembczy{\'{n}}ski~et~al.~\cite{dembczynski2010} in the context of single-label classification. However, in the multi-label setting, we consider the head as a vector, rather than a scalar, to enable rules to predict for several labels.

\subsection{Multi-label Loss Functions}
\label{sec:losses}

Various measures for evaluating the predictions provided by a multi-label classifier are commonly used in the literature (see, e.g., \cite{tsoumakas2010} for an overview). We focus on measures that assess the quality of predictions for $\numex$ examples and $\numlabels$ labels in terms of a single score $\measure ( \truelabelmatr, \predmatr ) \in \mathbb{R}_{+}$, where $\truelabelmatr, \predmatr \in \left \{ -1, +1 \right \}^{\numex \times \numlabels}$ are matrices that represent the true labels according to the ground truth, respectively the predicted labels provided by a classifier.

\subsubsection{Selected Evaluation Measures}

The \emph{Hamming loss} measures the fraction of incorrectly predicted labels among all labels and is defined as
\begin{equation}
\label{eq:hamming_loss}
\measure_{\text{Hamm.}} \left( \truelabelmatr, \predmatr \right) \coloneqq \frac{1}{\numex \numlabels} \sum_{\iterex = 1}^{\numex} \sum_{\iterlabels = 1}^{\numlabels} \, \llbracket \truelabel_{\iterex \iterlabels} \neq \predlabel_{\iterex \iterlabels} \rrbracket \, ,
\end{equation}
where $\llbracket P \rrbracket = 1$ if the predicate $P$ is true $= 0$ otherwise.

In addition, we use the \emph{subset 0/1 loss} to measure the fraction of examples for which at least one label is predicted incorrectly. Is it formally defined as
\begin{equation}
\label{eq:subset_01_loss}
\measure_{\text{subs.}} \left( \truelabelmatr, \predmatr \right) \coloneqq \frac{1}{\numex} \sum_{\iterex = 1}^{\numex} \llbracket \truelabelvect_{\iterex} \neq \predvect_{\iterex} \rrbracket \, .
\end{equation}
Both, the Hamming loss and the subset 0/1 loss, can be considered as generalizations of the 0/1 loss known from binary classification. Nevertheless, as will be discussed in the following \sect~\ref{sec:label_dependencies}, they have very different characteristics.

\subsubsection{Surrogate Loss Functions}

As seen in \sect~\ref{sec:ensembles}, the members of an ensemble predict vectors of numerical confidence scores, rather than binary label vectors. For this reason, discrete functions, such as the bipartition measures introduced above, are not suited to assess the quality of potential ensemble members during training. Instead, continuous loss functions that can be minimized in place of the actual target measure ought to be used as surrogates. For this purpose, we use multivariate (instead of univariate) loss functions $\loss: \{-1, +1\}^{\numlabels} \times \mathbb{R}^{\numlabels} \to \mathbb{R}_{+}$, which take two vectors $\truelabelvect_{\iterex}$ and $\confvect_{\iterex}$ as arguments. The former represents the true labeling of an example $\ex_{\iterex}$, whereas the latter corresponds to the predictions of the ensemble members according to \eqref{eq:confvect_aggregated}.

As surrogates for the Hamming loss and the subset 0/1 loss, we use different variants of the \emph{logistic loss}. This loss function, which is equivalent to \emph{cross-entropy}, is the basis for logistic regression and is commonly used in boosting approaches to single-label classification (early uses go back to Friedman~et~al.~\cite{friedman2000}). To cater for the characteristics of the Hamming loss, the \emph{label-wise logistic loss} applies the logistic loss function to each label individually:
\begin{equation}
\label{eq:log_loss_label_wise}
\loss_{\text{l.w.-log}} \left( \truelabelvect_{\iterex}, \confvect_{\iterex} \right) \coloneqq \sum_{\iterlabels = 1}^{\numlabels} \log \left( 1 + \exp \left( -\truelabel_{\iterex \iterlabels} \conf_{\iterex \iterlabels} \right) \right) \, .
\end{equation}
Following the formulation of this objective, $\conf_{\iterex \iterlabels}$ can be considered as log-odds, which estimates the probability of $\truelabel_{\iterex \iterlabels} = 1$ as $\logistic \left( \conf_{\iterex \iterlabels} \right) = \frac{1}{1 + \exp \left( -\conf_{\iterex \iterlabels} \right)}$. Under the assumption of label independence, the logistic loss has been shown to be a consistent surrogate loss for the Hamming loss \cite{dembczynski2012b,gao2013}.

As the label-wise logistic loss can be calculated by aggregating the values that result from applying the loss function to each label individually, it is label-wise decomposable. In contrast, the \emph{example-wise logistic loss}
\begin{equation}
\label{eq:log_loss_example_wise}
\loss_{\text{ex.w.-log}} \left( \truelabelvect_{\iterex}, \confvect_{\iterex} \right) \coloneqq \log \left( 1 + \sum_{\iterlabels = 1}^{\numlabels} \exp \left( -\truelabel_{\iterex \iterlabels} \conf_{\iterex \iterlabels} \right) \right)
\end{equation}
is non-decomposable, as it cannot be computed via label-wise aggregation. This smooth and convex surrogate is proposed by Amit~et~al.~\cite{amit2007}, who show that it provides an upper bound of the subset 0/1 loss.

\subsection{Label Dependence and (Non-)Decomposability}
\label{sec:label_dependencies}

The idea of modeling correlations between labels to improve the predictive performance of multi-label classifiers has been a driving force for research in MLC for many years. However, Dembczy{\'{n}}ski~et~al.~\cite{dembczynski2012a} brought up strong theoretical and empirical arguments that the type of loss function to be minimized, as well as the type of dependencies that occur in the data, strongly influence to what extent the exploitation of label dependencies may result in an improvement. The authors distinguish between two types of dependencies, namely \emph{marginal (unconditional)} and \emph{conditional dependence}. While the former refers to a lack of (stochastic) independence properties of the joint probability distribution $p \left( \truelabelvect \right) = p \left( \truelabel_{1}, \dots, \truelabel_{\numlabels} \right)$ on labelings $\truelabelvect$, the latter concerns the conditional probabilities $p \left( \truelabelvect \, | \, \ex \right)$, i.e., the distribution of labelings conditioned on an instance $\ex$.

According to the notion given in \sect~\ref{sec:rules}, the rules we aim to learn contribute to the final prediction of labels for which they predict a non-zero confidence score and abstain otherwise. The head of a \emph{multi-label rule} contains multiple non-zero scores, which enables one to express conditional dependencies between the corresponding labels, where the rule's body is tailored to cover a region of the attribute space where these dependencies hold. In contrast, \emph{single-label rules} are tailored to exactly one label and ignore the others, for which reason they are unable to explicitly express conditional dependencies.

As discussed in \sect~\ref{sec:losses}, we are interested in the Hamming loss and the subset 0/1 loss, which are representatives for decomposable and non-de\-com\-pos\-able loss functions, respectively. In the case of decomposability, modeling dependencies between the labels cannot be expected to drastically improve predictive performance \cite{dembczynski2012a}. For this reason, we expect that single-label rules suffice for minimizing Hamming loss on most data sets. In contrast, given that the labels in a data set are not conditionally independent, the ability to model dependencies is required to effectively minimize non-decomposable losses. Hence, we expect that the ability to learn multi-label rules is crucial for minimizing the subset 0/1 loss.

\section{Gradient Boosting using Multivariate Loss Functions}
\label{sec:boosting}

As our first contribution, we formulate an extension of the gradient boosting framework to multivariate loss functions. This formulation, which should be flexible enough to use any decomposable or non-decomposable loss function, as long as it is differentiable, serves as the basis of the MLC method that is proposed in \sect~\ref{sec:algorithm} as the main contribution of this paper.

\subsection{Stagewise Additive Modeling}
\label{sec:stagewise_modeling}

We aim at learning an ensemble of additive functions $\ensemble = \{ \classifier_{1}, \dots, \classifier_{\numclassifiers} \}$ as introduced in \sect~\ref{sec:ensembles}. It should be trained in a way such that the expected empirical risk with respect to a certain (surrogate) loss function $\loss$ is minimized. Thus, we are concerned with minimizing the regularized training objective
\begin{equation}
\label{eq:objective}
\objective \left( \ensemble \right) = \sum_{\iterex = 1}^{\numex} \loss \left( \truelabelvect_{\iterex}, \confvect_{\iterex} \right) + \sum_{\iterclassifiers = 1}^{\numclassifiers} \regterm \left( \classifier_{\iterclassifiers} \right) \, ,
\end{equation}
where $\regterm$ denotes an (optional) regularization term that may be used to penalize the complexity of the individual ensemble members to avoid overfitting and to ensure the convergence towards a global optimum if $\loss$ is not convex.

Unfortunately, constructing an ensemble of additive functions that minimizes the objective given above is a hard optimization problem. In gradient boosting, this problem is tackled by training the model in a stagewise procedure, where the individual ensemble members are added one after the other, as originally proposed by Friedman~et~al.~\cite{friedman2000}. At each iteration $\iterclassifiers$, the vector of scores $\ensemble_{\iterclassifiers} \left( \ex_{\iterex} \right)$ that is predicted by the existing ensemble members for an example $\ex_{\iterex}$ can be calculated based on the predictions of the previous iteration:
\begin{equation}
\label{eq:stagewise_prediction}
\ensemble_{\iterclassifiers} \left( \ex_{\iterex} \right) = \ensemble_{\iterclassifiers - 1} \left( \ex_{\iterex} \right) + \classifier_{\iterclassifiers} \left( \ex_{\iterex} \right) = \left( \confvect_{\iterex}^{1} + \dots + \confvect_{\iterex}^{\iterclassifiers - 1} \right) + \confvect_{\iterex}^{\iterclassifiers} \, .
\end{equation}
Substituting the additive calculation of the predictions into the objective function given in \eqref{eq:objective} yields the following objective to be minimized by the ensemble member that is added in the $\iterclassifiers$-th iteration:
\begin{equation}
\label{eq:stagewise_objective}
\objective \left( \classifier_{\iterclassifiers} \right) = \sum_{\iterex = 1}^{\numex} \loss \left( \truelabelvect_{\iterex}, \ensemble_{\iterclassifiers - 1} \left( \ex_{\iterex} \right) + \confvect_{\iterex}^{\iterclassifiers} \right) + \regterm \left( \classifier_{\iterclassifiers} \right) \, .
\end{equation}

\subsection{Multivariate Taylor Approximation}
\label{sec:taylor_approximation}

To be able to efficiently minimize the training objective when adding a new ensemble member $\classifier_{\iterclassifiers}$, we rewrite \eqref{eq:stagewise_objective} in terms of the second-order multivariate Taylor approximation
\begin{equation}
\label{eq:taylor_approximation}
\objective \left( \classifier_{\iterclassifiers} \right) \approx \sum_{\iterex = 1}^{\numex} \left( \loss \left( \truelabelvect_{\iterex},  \ensemble_{\iterclassifiers - 1} \left( \ex_{\iterex} \right) \right) + \gradientvect_{\iterex} \confvect_{\iterex}^{\iterclassifiers} + \frac{1}{2} \confvect_{\iterex}^{\iterclassifiers} \hessianmatr_{\iterex} \confvect_{\iterex}^{\iterclassifiers} \right) + \regterm \left( \classifier_{\iterclassifiers} \right) \, ,
\end{equation}
where $\gradientvect_{\iterex} = \left( \gradient_{\iterex 1}, \dots, \gradient_{\iterex \numlabels} \right)$ denotes the vector of first-order partial derivatives of the loss function $\loss$ with respect to the existing ensemble members' predictions for a particular example $\ex_{\iterex}$ and individual labels $\lambda_{\iterlabels}$. Accordingly, the Hessian matrix $\hessianmatr_{\iterex} = \left( \left( \hessian_{\iterex 1 1} \dots \hessian_{\iterex 1 \numlabels} \right), \dots, \left( \hessian_{\iterex \numlabels 1} \dots \hessian_{\iterex \numlabels \numlabels} \right) \right)$ consists of all second-order partial derivatives:
\begin{equation}
\label{eq:gradients_hessians}
\gradient_{\iterex \iterrows} = \frac{\partial \loss}{\partial \conf_{\iterex \iterrows}} \left( \truelabelvect_{\iterex},  \ensemble_{\iterclassifiers - 1} \left( \ex_{\iterex} \right) \right), \quad \hessian_{\iterex \iterrows \itercols} = \frac{\partial \loss}{\partial \conf_{\iterex \iterrows} \partial \conf_{\iterex \itercols}} \left( \truelabelvect_{\iterex},  \ensemble_{\iterclassifiers - 1} \left( \ex_{\iterex} \right) \right) \, .
\end{equation}
By removing constant terms, \eqref{eq:taylor_approximation} can be further simplified, resulting in the approximated training objective
\begin{equation}
\label{eq:stagewise_objective_approx}
\approxobjective \left( \classifier_{\iterclassifiers} \right) = \sum_{\iterex = 1}^{\numex} \left( \gradientvect_{\iterex} \confvect_{\iterex}^{\iterclassifiers} + \frac{1}{2} \confvect_{\iterex}^{\iterclassifiers} \hessianmatr_{\iterex} \confvect_{\iterex}^{\iterclassifiers} \right) + \regterm \left( \classifier_{\iterclassifiers} \right) \, .
\end{equation}
In each training iteration, the objective function $\approxobjective$ can be used as a quality measure to decide which of the potential ensemble members improves the current model the most. This requires the predictions of the potential ensemble members for examples $\ex_{\iterex}$ to be known. How to find these predictions depends on the type of ensemble members used and the loss function at hand. In \sect~\ref{sec:algorithm}, we present solutions to this problem, using classification rules as ensemble members.

\section{Learning Boosted Multi-label Classification Rules}
\label{sec:algorithm}

Based on the general framework defined in the previous section, we now present BOOMER, a concrete stagewise algorithm for learning an ensemble of gradient boosted single- or multi-label rules $\ensemble = \{ \rul_{1}, \dots, \rul_{\numclassifiers} \}$ that minimizes a given loss function in expectation\footnote{An implementation is available at \url{https://www.github.com/mrapp-ke/Boomer}}. The basic algorithm is outlined in \alg~\ref{alg:train}.

\begin{algorithm}[t!]
\SetKwInOut{Input}{input}\SetKwInOut{Output}{output}
  \Input{Training examples $\dataset = \left \{ \left( \ex_{\iterex}, \truelabelvect_{\iterex} \right) \right \}_{\iterex}^{\numex}$, first and second derivative $\loss'$ and $\loss''$ of the loss function, number of rules $\numclassifiers$, shrinkage parameter $\shrinkage$}
  \Output{Ensemble of rules $\ensemble$}
  $\gradientset = \left \{ \gradientvect_{\iterex} \right \}_{\iterex}^{\numex}, \hessianset = \left \{ \hessianmatr_{\iterex} \right \}_{\iterex}^{\numex} =$ calculate gradients and Hessians w.r.t. $\loss'$ and $\loss''$ \\ \label{ln:init_gradients}
  $\rul_{1}: \body_{1} \rightarrow \head_{1}$ with $\body_{1} \left( \ex \right) = 1, \forall \ex$ and $\head_{1} = \textsc{find\_head} \left( \dataset, \gradientset, \hessianset, \body_{1} \right)$ \hfill \comment{\sect~\ref{sec:find_head}} \\
  \For{$\iterclassifiers = 2$ \KwTo $\numclassifiers$}{
    $\gradientset$, $\hessianset =$ update gradients and Hessians of examples covered by $\rul_{\iterclassifiers - 1}$ \\ \label{ln:update_gradients}
    $\dataset' =$ randomly draw $\numex$ examples from $\dataset$ (with replacement) \\ \label{ln:bagging}
    $\rul_{\iterclassifiers}: \body_{\iterclassifiers} \rightarrow \head_{\iterclassifiers} = \textsc{refine\_rule} \left( \dataset', \gradientset, \hessianset \right)$ \hfill \comment{\sect~\ref{sec:refine_rule}} \\
    $\head_{\iterclassifiers} = \textsc{find\_head} \left( \dataset, \gradientset, \hessianset, \body_{\iterclassifiers} \right)$ \\ \label{ln:recompute_head}
    $\head_{\iterclassifiers} = \shrinkage \cdot \head_{\iterclassifiers}$ \label{ln:shrinkage}
  }
  \Return ensemble of rules $\ensemble = \{ \rul_{1}, \dots, \rul_{\numclassifiers} \}$
  \caption{Learning an ensemble of boosted classification rules}
  \label{alg:train}
\end{algorithm}

Because rules, unlike other classifiers like e.g. decision trees, only provide predictions for examples they cover, the first rule $\rul_{1}$ in the ensemble is a \emph{default rule}, which covers all examples, i.e., $\body_{1} \left( \ex \right) = 1, \forall \ex \in \attrspace$. In subsequent iterations, more specific rules are added. All rules, including the default rule, contribute to the final predictions of the ensemble according to their confidence scores, which are chosen such that the objective function in \eqref{eq:stagewise_objective_approx} is minimized. In each iteration $\iterclassifiers$, this requires the gradients and Hessians to be (re-)calculated based on the confidence scores $\conf_{\iterex \iterlabels}$ that are predicted for each example $\ex_{\iterex}$ and label $\labl_{\iterlabels}$ by the current model ($\conf_{\iterex \iterlabels} = 0, \forall \iterex, \iterlabels$ if $t = 1$), as well as the true labels $\truelabel_{\iterex \iterlabels} \in \left \{ -1, +1 \right \}$ (cf.~\alg~\ref{alg:train}, \alglns~\ref{ln:init_gradients} and \ref{ln:update_gradients}). While the default rule always provides a confidence score for each label, all of the remaining rules may either predict for a single label or for all labels, depending on a hyper-parameter. We consider both variants in the experimental study presented in \sect~\ref{sec:evaluation}. The computations necessary to obtain loss-minimizing predictions for the default rule and each of the remaining rules are presented in \sect~\ref{sec:find_head}.

To learn the rules $\rul_{2}, \dots, \rul_{\numclassifiers}$, we use a greedy procedure where the body is iteratively refined by adding new conditions and the head is adjusted accordingly in each step. The algorithm used for rule refinement is discussed in detail in \sect~\ref{sec:refine_rule}. To reduce the variance of the ensemble members, each rule is learned on a different subsample of the training examples, randomly drawn with replacement as in \emph{bagging}, which results in more diversified and less correlated rules. However, once a rule has been learned, we recompute its predictions on the entire training data (cf.~\alg~\ref{alg:train}, \algln~\ref{ln:recompute_head}), which we have found to effectively prevent overfitting the sub-sample used for learning the rule. As an additional measure to reduce the risk of fitting noise in the data, the scores predicted by a rule may be multiplied by a \emph{shrinkage} parameter $\shrinkage \in \left( 0, 1\right]$ (cf.~\alg~\ref{alg:train}, \algln~\ref{ln:shrinkage}). Small values for $\shrinkage$, which can be considered as the learning rate, reduce the impact of individual rules on the model \cite{hastie2009}.

\subsection{Computation of Loss-minimizing Scores}
\label{sec:find_head}

As illustrated in \alg~\ref{alg:find_head}, the function \textsc{find\_head} is used to find optimal confidence scores to be predicted by a particular rule $\rul$, i.e., scores that minimize the objective function $\approxobjective$ introduced in \eqref{eq:stagewise_objective_approx}. Because of the fact that rules provide the same predictions $\head$ for all examples $\ex_{\iterex}$ they cover and abstain for the others, the objective function can be further simplified. As the addition is commutative, we can sum up the gradient vectors and Hessian matrices that correspond to the covered examples (cf.~\alg~\ref{alg:find_head}, \algln~\ref{ln:gradient_sum}), resulting in the objective
\begin{equation}
\label{eq:rule_objective}
\approxobjective \left( \rul_{\iterclassifiers} \right) = \gradientvect \head + \frac{1}{2} \head \hessianmatr \head + \regterm \left( \rul_{\iterclassifiers} \right) \, ,
\end{equation}
where $\gradientvect = \sum_{\iterex} \body \left( \ex_{\iterex} \right) \gradientvect_{\iterex}$ denotes the element-wise sum of the gradient vectors and $\hessianmatr = \sum_{\iterex} \body \left( \ex_{\iterex} \right) \hessianmatr_{\iterex}$ corresponds to the sum of the Hessian matrices.

To penalize extreme predictions, we use the $L_2$ regularization term
\begin{equation}
\label{eq:regterm}
\regterm_{\text{L2}} \left( \rul_{\iterclassifiers} \right) = \frac{1}{2} \regweight \left \lVert \confvect^{\iterclassifiers} \right \rVert_{2}^{2} \, ,
\end{equation}
where $\left \lVert \vec{x} \right \rVert_{2}$ denotes the Euclidean norm and $\regweight \geq 0$ is the regularization weight.

\begin{algorithm}[t!]
\SetKwInOut{Input}{input}\SetKwInOut{Output}{output}
  \Input{(Sub-sample of) training examples $\dataset = \left \{ \left( \ex_{\iterex}, \truelabelvect_{\iterex} \right) \right \}_{\iterex}^{\numex}$, \\
  gradients $\gradientset = \left \{ \gradientvect_{\iterex} \right \}_{\iterex}^{\numex}$, Hessians $\hessianset = \left \{ \hessianmatr_{\iterex} \right \}_{\iterex}^{\numex}$, body $\body$}
  \Output{Single- or multi-label head $\head$}
  $\gradientvect = \sum_{\iterex} \body \left( \ex_{\iterex} \right) \gradientvect_{\iterex}, \hessianmatr = \sum_{\iterex} \body \left( \ex_{\iterex} \right) \hessianmatr_{\iterex}$ \\ \label{ln:gradient_sum}
  \If{\upshape loss function is decomposable \textbf{or} searching for a single-label head}{
    $\head =$ obtain $\conf_{\iterlabels}$ w.r.t. $\gradientvect$ and $\hessianmatr$ for each label independently acc. to \eqref{eq:optimal_score_decomposable} \\
    \If{\upshape searching for a single-label head}{
      $\head =$ find best single-label prediction $\conf_{\iterlabels} \in \head$ w.r.t. \eqref{eq:stagewise_objective_approx} 
    }
  }
  \Else{
    $\head =$ obtain $\left( \conf_{1}, \dots, \conf_{\numlabels} \right)$ w.r.t. $\gradientvect$ and $\hessianmatr$ by solving the linear system in \eqref{eq:optimal_score_non_decomposable}
  }
  \Return head $\head$
  \caption{\textsc{find\_head}}
  \label{alg:find_head}
\end{algorithm}

To ensure that the predictions $\confvect$ minimize the regularized training objective $\approxobjective$, we equate the first partial derivative of \eqref{eq:rule_objective} with respect to $\confvect$ with zero:
\begin{align}
\begin{split}
\label{eq:optimal_score_non_decomposable}
\frac{\partial \approxobjective}{\partial \confvect} \left( \rul_{\iterclassifiers} \right) = & \gradientvect + \hessianmatr \confvect + \regweight \confvect = \gradientvect + \left( \hessianmatr + \diag \left( \regweight \right) \right) \confvect = 0 \\
\Longleftrightarrow & \left( \hessianmatr + \diag \left( \regweight \right) \right) \confvect = -\gradientvect \, ,
\end{split}
\end{align}
where $\diag \left( \regweight \right)$ is a diagonal matrix with $\regweight$ on the diagonal.

\eqref{eq:optimal_score_non_decomposable} can be considered as a system of $\numlabels$ linear equations, where $\hessianmatr + \diag \left( \regweight \right)$ is a matrix of coefficients, $-\gradientvect$ is a vector of ordinates and $\confvect$ is the vector of unknowns to be determined. For commonly used loss functions, including the ones in \sect~\ref{sec:losses}, the sums of Hessians $\hessian_{\iterrows \itercols}$ and $\hessian_{\itercols \iterrows}$ are equal. Consequently, the matrix of coefficients is symmetrical.

In the general case, i.e., if the loss function is non-decomposable, the linear system in \eqref{eq:optimal_score_non_decomposable} must be solved to determine the optimal multi-label head $\head$. However, when dealing with a decomposable loss function, the first and second derivative with respect to a particular element $\conf_{\iterrows} \in \confvect$ is independent of any other element $\conf_{\itercols} \in \confvect$. This causes the sums of Hessians $\hessian_{\iterrows \itercols}$ that do not exclusively depend on $\conf_{\iterrows}$, i.e., if $\iterrows \neq \itercols$, to become zero. In such case, the linear system reduces to $\numlabels$ independent equations, one for each label. This enables to compute the optimal prediction $\conf_{\iterrows}$ for the $\iterrows$-th label as
\begin{equation}
\label{eq:optimal_score_decomposable}
\conf_{\iterrows} = -\frac{\gradient_{\iterrows}}{\hessian_{\iterrows \iterrows} + \regweight} \, .
\end{equation}
Similarly, when dealing with single-label rules that predict for the $\iterrows$-th label, the predictions $\conf_{\itercols}$ with $\itercols \neq \iterrows$ are known to be zero, because the rule will abstain for the corresponding labels. Consequently, \eqref{eq:optimal_score_decomposable} can be used to determine the predictions of single-label rules even if the loss function is non-decomposable.

\subsection{Refinement of Rules}
\label{sec:refine_rule}

To learn a new rule, we use a top-down greedy search, commonly used in inductive rule learning (see, e.g., \cite{fuernkranz2012} for an overview). \alg~\ref{alg:refine_rule} is meant to outline the general procedure and does not include any algorithmic optimizations that can drastically improve the computational efficiency in practice.

The search starts with an empty body that is successively refined by adding additional conditions. Adding conditions to its body causes the rule to become more specific and results in less examples being covered. The conditions, which may be used to refine an existing body, result from the values of the covered examples in case of nominal attributes or from averaging adjacent values in case of numerical attributes. In addition to bagging, we use \emph{random feature selection}, as in random forests, to ensure that the rules in an ensemble are more diverse (cf.~\alg~\ref{alg:refine_rule}, \algln~\ref{ln:random_feature_selection}). For each condition that may be added to the current body at a particular iteration, the head of the rule is updated via the function \textsc{find\_head} that has already been discussed in \sect~\ref{sec:find_head} (cf.~\alg~\ref{alg:refine_rule}, \algln~\ref{ln:update_head}). As a restriction, in the case of single-label rules, each refinement of the current rule must predict for the same label (omitted in \alg~\ref{alg:refine_rule} for brevity). Among all refinements, the one that minimizes the regularized objective in \eqref{eq:rule_objective} is chosen. If no refinement results in an improvement according to said objective, the refinement process~stops. We do not use any additional stopping criteria.

\begin{algorithm}[t!]
\SetKwInOut{Input}{input}\SetKwInOut{Output}{output}
  \Input{(Sub-sample of) training examples $\dataset = \left \{ \left( \ex_{\iterex}, \truelabelvect_{\iterex} \right) \right \}_{\iterex}^{\numex}$, \\ 
  gradients $\gradientset = \left \{ \gradientvect_{\iterex} \right \}_{\iterex}^{\numex}$, Hessians $\hessianset = \left \{ \hessianmatr_{\iterex} \right \}_{\iterex}^{\numex}$, current rule $\rul$ (optional)}
  \Output{Best rule $\rul^{*}$}
  $\rul^{*} = \rul$ \\
  $A' =$ randomly select $\lfloor \log_2 \left( \numattr -1 \right) + 1 \rfloor$ out of $\numattr$ attributes from $\dataset$ \\ \label{ln:random_feature_selection}
  \ForEach{\upshape possible condition $\cond$ on attributes $A'$ and examples $\dataset$}{
    $\rul': \body' \rightarrow \head' =$ copy of current rule $\rul$ \\
    add condition $\cond$ to body $\body'$ \\
    $\head' = \textsc{find\_head} \left( \dataset, \gradientset, \hessianset, \body' \right)$ \hfill \comment{\sect~\ref{sec:find_head}} \\ \label{ln:update_head}
    \If{\upshape $\approxobjective \left( \rul' \right) < \approxobjective \left( \rul^{*} \right)$ w.r.t. $\gradientset$ and $\hessianset$}{
      $\rul^{*} = \rul'$ \\
    }
  }
  \If{$\rul^{*} \neq \rul$}{
    $\dataset' =$ subset of $\dataset$ covered by $\rul^{*}$ \\
    \Return $\textsc{refine\_rule} \left( \dataset', \gradientset, \hessianset, \rul^{*} \right)$
  }
  \Return best rule $\rul^{*}$ \\
  \caption{\textsc{refine\_rule}}
  \label{alg:refine_rule}
\end{algorithm}

\subsection{Prediction}
\label{sec:prediction}

Predicting for an example $\ex_{\iterex}$ involves two steps: First, the scores provided by individual rules are aggregated into a vector of confidence scores $\confvect_{\iterex}$ according to \eqref{eq:confvect_aggregated}. Second, $\confvect_{\iterex}$ must be turned into a binary label vector $\predvect_{\iterex} \in \labelspace$. As it should minimize the expected risk with respect to the used loss function, i.e., $\predvect_{\iterex} = \argmin_{\predvect_{\iterex}} \loss \left( \truelabelvect_{\iterex}, \predvect_{\iterex} \right)$, it should be chosen in a way that is tailored to the loss function at hand. 

In case of the label-wise logistic loss in \eqref{eq:log_loss_label_wise}, we compute the prediction as
\begin{equation}
\label{eq:prediction_label_wise}
\predvect_{\iterex} = \left( \sgn \left( \conf_{\iterex 1} \right), \dots, \sgn \left( \conf_{\iterex \numlabels} \right) \right) \, ,
\end{equation}
where $\sgn(x) = +1$ if $x > 0$ and $= -1$ otherwise.

To predict the label vector that minimizes the example-wise logistic loss given in \eqref{eq:log_loss_example_wise}, we return the label vector among the vectors in the training data, which minimizes the loss, i.e.,
\begin{equation}
\label{eq:prediction_example_wise}
\predvect_{\iterex} = \argmin_{\truelabelvect \in \dataset} \loss_{\text{ex.w.-log}} \left( \truelabelvect, \confvect_{\iterex} \right) \,.
\end{equation}
The main reason for picking only a known label vector was to ensure a fair comparison to the main competitor for non-decomposable losses, LP, which is also restricted to returning only label vectors seen in the training data. Moreover, Senge et al.~\cite{senge2013} argue that often only a small fraction of the $2^{\numlabels}$ possible label subsets are observed in practice, for which reason the correctness of an unseen label combination becomes increasingly unlikely for large data sets, and, in fact, our preliminary experiments confirmed this.

\section{Evaluation}
\label{sec:evaluation}

We evaluate the ability of our approach to minimize Hamming and subset 0/1 loss, using the label-wise and example-wise logistic loss as surrogates. In each case, we consider both, single- and multi-label rules, resulting in four different variants (\textsl{l.w.-log.~single}, \textsl{l.w.-log.~multi}, \textsl{ex.w.-log.~single}, and \textsl{ex.w.-log.~multi}). 
For each of them, we tune the shrinkage parameter $\shrinkage \in \left \{ 0.1, 0.3, 0.5 \right \}$ and the regularization weight $\regweight \in \left \{ 0.0, 0.25, 1.0, 4.0, 16.0, 64.0 \right \}$, using  \emph{Bootstrap Bias Corrected Cross Validation (BBC-CV)} \cite{tsamardinos2018}, which allows to incorporate parameter tuning and evaluation in a computationally efficient manner.

\subsection{Synthetic Data}

\begin{figure}[t!]
\centering
\begin{subfigure}[b]{0.475\textwidth}
  \centering
  \includegraphics[width=\textwidth]{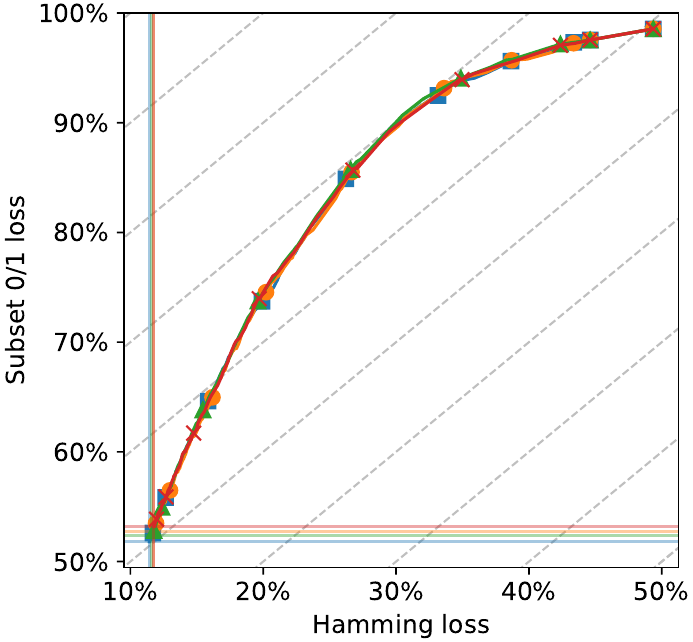}
  \caption{Marginal independence}
  \label{fig:synthetic_marginal_independence}
\end{subfigure}
\hfill
\begin{subfigure}[b]{0.475\textwidth}
  \centering 
  \includegraphics[width=\textwidth]{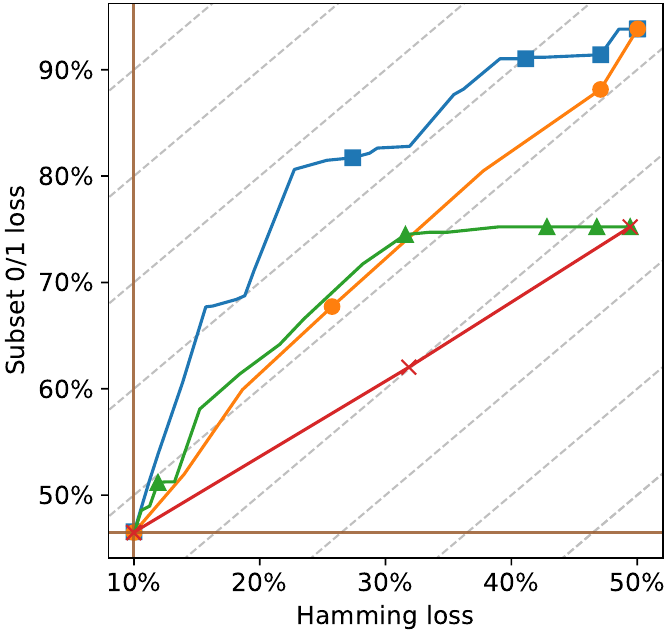}
  \caption{Strong marginal dependence}    
  \label{fig:synthetic_marginal_dependence}
\end{subfigure}
\vskip\baselineskip
\begin{subfigure}[b]{0.475\textwidth}   
  \centering 
  \includegraphics[width=\textwidth]{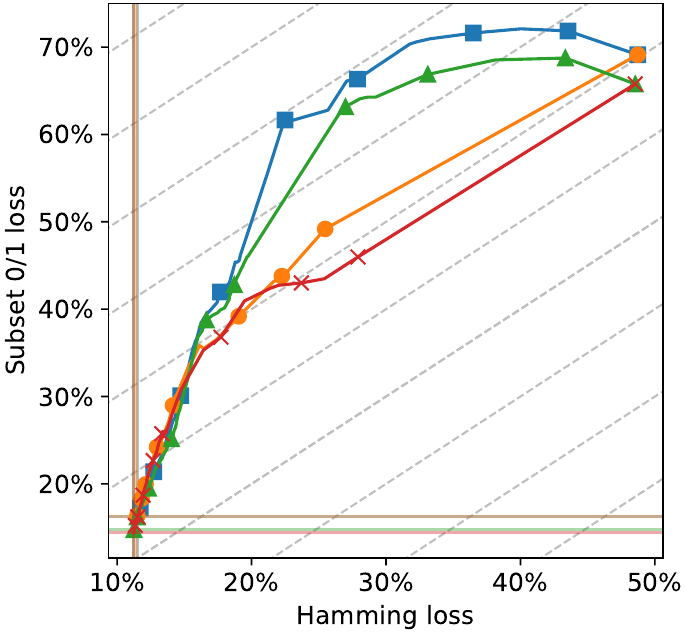}
  \caption{Strong conditional dependence}    
  \label{fig:synthetic_conditional_dependence}
\end{subfigure}
\hfill
\begin{minipage}[b]{0.475\textwidth}
  \centering
  \includegraphics[width=\textwidth]{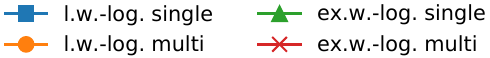}
  \vskip\baselineskip
  \caption{Predictive performance of different approaches with respect to the Hamming loss and the subset 0/1 loss on three synthetic data sets. Starting at the top right, each curve shows the performance as more rules are added. The slope of the isometrics refers to an even improvement of Hamming and subset 0/1 loss. The symbols indicate $1, 2, 4, 8, \dots, 512, 1000$ rules in the model.}
  \label{fig:synthetic}
\end{minipage}
\end{figure}

We use three synthetic data sets, each containing $10,000$ examples and six labels, as proposed by Dembczy\'nski et al. \cite{dembczynski2012b} for analyzing the effects of using single- or multi-label rules with respect to different types of label dependencies. The attributes correspond to two-dimensional points drawn randomly from the unit circle and the labels are assigned according to linear decision boundaries through its origin. The results are shown in \fig~\ref{fig:synthetic}.

In the case of \emph{marginal independence}, points are sampled for each label independently. Noise is introduced by randomly flipping 10\% of the labels, such that the scores that are achievable by the Bayes-optimal classifier on an independent test set are $10\%$ for Hamming loss and $\approx 47\%$ for subset 0/1 loss. As it is not possible --- by construction of the data set --- to find a rule body that is suited to predict for more than one label, multi-label rules cannot be expected to provide any advantage. 
In fact, despite very similar trajectories, the single-label variants achieve slightly better losses in the limit. This indicates that, most probably due to the number of uninformative features, the approaches that aim at learning multi-label rules struggle to induce pure single-label rules.

For modeling a \emph{strong marginal dependence} between the labels, a small angle between the linear boundaries of two labels is chosen, so that they mostly coincide for the respective examples. Starting from different default rules, depending on the loss function they minimize, the algorithms converge to the same performances, which indicates that all variants can similarly deal with marginal dependencies. However, when using multi-label rules, the final subset 0/1 score is approached in a faster and more steady way, as a single rule provides predictions for several labels at once. In fact, it is remarkable that the ex.w.-log. multi variants already converge after two rules, whereas ex.w.-log. single needs six, one for each label, and optimizing for label-wise logarithmic loss takes much longer.

Finally, \emph{strong conditional dependence} is obtained by randomly switching all labels for $10\%$ of the examples. As a result, the score that is achievable by the Bayes-optimal classifier is $10\%$ for both losses. While the trajectories are similar to before, the variants that optimize the non-decomposable loss achieve better results in the limit. Unlike the approaches that consider the labels independently, they seem to be less prone to the noise that is introduced at the example-level.

\subsection{Real-World Benchmark Data}

\begin{table}[t]
\caption{Predictive performance (in percent) of different approaches with respect to Hamming loss, subset 0/1 loss and the example-based F1-measure. For each evaluation measure and data set, we report the ranks of the different approaches (small numbers) and highlight the best approach (bold text).}
\label{tab:evaluation_benchmark}
\centering
\begin{tabular}{|l|l||r c r c|r c r c||r c r c r c|}
\cline{3-16}
\multicolumn{2}{c|}{}     
 & \multicolumn{4}{c|}{l.w.-log.}
 & \multicolumn{4}{c||}{ex.w.-log.}
 & \multicolumn{6}{c|}{XGBoost} \\
\multicolumn{2}{c|}{}     
 & \multicolumn{2}{c}{single}
 & \multicolumn{2}{c|}{multi}
 & \multicolumn{2}{c}{single}
 & \multicolumn{2}{c||}{multi}
 & \multicolumn{2}{c}{BR}
 & \multicolumn{2}{c}{LP}
 & \multicolumn{2}{c|}{CC} \\
\cline{3-16}
\noalign{\vskip\doublerulesep}
\cline{1-16}
\parbox[c]{3mm}{\multirow{9}{*}{\rotatebox[origin=c]{90}{Subset 0/1 loss}}}
 & \textsc{birds}
 & $38.72$              & $\scriptstyle 2$
 & $40.43$              & $\scriptstyle 6$
 & $39.57$              & $\scriptstyle 4.5$
 & $39.15$              & $\scriptstyle 3$
 & $39.57$              & $\scriptstyle 4.5$
 & $40.85$              & $\scriptstyle 7$
 & $\boldsymbol{38.30}$ & $\scriptstyle 1$ \\
 & \textsc{emotions}
 & $73.33$              & $\scriptstyle 6.5$
 & $73.33$              & $\scriptstyle 6.5$
 & $70.48$              & $\scriptstyle 3.5$
 & $65.24$              & $\scriptstyle 2$
 & $70.48$              & $\scriptstyle 3.5$
 & $\boldsymbol{60.00}$ & $\scriptstyle 1$
 & $71.43$              & $\scriptstyle 5$ \\
 & \textsc{enron}
 & $87.81$              & $\scriptstyle 7$
 & $86.82$              & $\scriptstyle 6$
 & $84.35$              & $\scriptstyle 4$
 & $83.53$              & $\scriptstyle 2$
 & $85.34$              & $\scriptstyle 5$
 & $\boldsymbol{82.70}$ & $\scriptstyle 1$
 & $83.86$              & $\scriptstyle 3$ \\
 & \textsc{llog}
 & $78.85$              & $\scriptstyle 5$
 & $78.85$              & $\scriptstyle 5$
 & $78.27$              & $\scriptstyle 3$
 & $76.35$              & $\scriptstyle 2$
 & $80.96$              & $\scriptstyle 7$
 & $\boldsymbol{70.38}$ & $\scriptstyle 1$
 & $78.85$              & $\scriptstyle 5$ \\
 & \textsc{medical}
 & $28.45$              & $\scriptstyle 3$
 & $30.16$              & $\scriptstyle 4$
 & $23.90$              & $\scriptstyle 2$
 & $\boldsymbol{23.04}$ & $\scriptstyle 1$
 & $56.90$              & $\scriptstyle 7$
 & $41.11$              & $\scriptstyle 5$
 & $44.95$              & $\scriptstyle 6$ \\
 & \textsc{scene}
 & $39.33$              & $\scriptstyle 7$
 & $33.93$              & $\scriptstyle 5$
 & $25.17$              & $\scriptstyle 3$
 & $\boldsymbol{23.26}$ & $\scriptstyle 1$
 & $34.72$              & $\scriptstyle 6$
 & $24.16$              & $\scriptstyle 2$
 & $30.11$              & $\scriptstyle 4$ \\
 & \textsc{slashdot}
 & $65.29$              & $\scriptstyle 5$
 & $62.89$              & $\scriptstyle 4$
 & $53.30$              & $\scriptstyle 3$
 & $\boldsymbol{49.75}$ & $\scriptstyle 1$
 & $72.04$              & $\scriptstyle 7$
 & $52.94$              & $\scriptstyle 2$
 & $66.96$              & $\scriptstyle 6$ \\
 & \textsc{yeast}
 & $83.72$              & $\scriptstyle 7$
 & $82.27$              & $\scriptstyle 5$
 & $78.60$              & $\scriptstyle 4$
 & $75.81$              & $\scriptstyle 2$
 & $82.72$              & $\scriptstyle 6$
 & $\boldsymbol{75.70}$ & $\scriptstyle 1$
 & $76.59$              & $\scriptstyle 3$ \\
 \cline{2-16}
 & Avg. rank
 & $5.31$               &
 & $5.19$               &
 & $3.38$               &
 & $\boldsymbol{1.75}$  &
 & $5.75$               &
 & $2.50$               &
 & $4.13$               & \\
\cline{1-16}
\noalign{\vskip\doublerulesep}
\cline{1-16}
\parbox[c]{3mm}{\multirow{9}{*}{\rotatebox[origin=c]{90}{Hamming loss}}}
 & \textsc{birds}
 & $3.52$               & $\scriptstyle 5$
 & $\boldsymbol{3.16}$  & $\scriptstyle 1$
 & $3.58$               & $\scriptstyle 6$
 & $3.23$               & $\scriptstyle 2$
 & $3.43$               & $\scriptstyle 3$
 & $4.03$               & $\scriptstyle 7$
 & $3.45$               & $\scriptstyle 4$ \\
 & \textsc{emotions}
 & $18.81$              & $\scriptstyle 5$
 & $20.00$              & $\scriptstyle 7$
 & $18.17$              & $\scriptstyle 2$
 & $18.41$              & $\scriptstyle 3$
 & $18.73$              & $\scriptstyle 4$
 & $\boldsymbol{18.02}$ & $\scriptstyle 1$
 & $19.29$              & $\scriptstyle 6$ \\
 & \textsc{enron}
 & $4.56$               & $\scriptstyle 3$
 & $\boldsymbol{4.49}$  & $\scriptstyle 1$
 & $4.90$               & $\scriptstyle 5$
 & $4.91$               & $\scriptstyle 6$
 & $4.52$               & $\scriptstyle 2$
 & $5.75$               & $\scriptstyle 7$
 & $4.63$               & $\scriptstyle 4$ \\
 & \textsc{llog}
 & $1.48$               & $\scriptstyle 2.5$
 & $1.48$               & $\scriptstyle 2.5$
 & $1.49$               & $\scriptstyle 4.5$
 & $\boldsymbol{1.45}$  & $\scriptstyle 1$
 & $1.49$               & $\scriptstyle 4.5$
 & $2.13$               & $\scriptstyle 7$
 & $1.60$               & $\scriptstyle 6$ \\
 & \textsc{medical}
 & $0.84$               & $\scriptstyle 2.5$
 & $0.86$               & $\scriptstyle 4$
 & $0.84$               & $\scriptstyle 2.5$
 & $\boldsymbol{0.79}$  & $\scriptstyle 1$
 & $1.69$               & $\scriptstyle 7$
 & $1.55$               & $\scriptstyle 6$
 & $1.36$               & $\scriptstyle 5$ \\
 & \textsc{scene}
 & $8.16$               & $\scriptstyle 7$
 & $7.42$               & $\scriptstyle 6$
 & $7.21$               & $\scriptstyle 4$
 & $\boldsymbol{6.63}$  & $\scriptstyle 1$
 & $7.19$               & $\scriptstyle 3$
 & $7.27$               & $\scriptstyle 5$
 & $6.85$               & $\scriptstyle 2$ \\
 & \textsc{slashdot}
 & $\boldsymbol{4.15}$  & $\scriptstyle 1$
 & $4.17$               & $\scriptstyle 2$
 & $5.03$               & $\scriptstyle 6$
 & $4.64$               & $\scriptstyle 5$
 & $4.61$               & $\scriptstyle 4$
 & $5.16$               & $\scriptstyle 7$
 & $4.54$               & $\scriptstyle 3$ \\
 & \textsc{yeast}
 & $19.27$              & $\scriptstyle 2$
 & $19.84$              & $\scriptstyle 5$
 & $19.44$              & $\scriptstyle 4$
 & $19.29$              & $\scriptstyle 3$
 & $\boldsymbol{19.13}$ & $\scriptstyle 1$
 & $21.22$              & $\scriptstyle 7$
 & $20.11$              & $\scriptstyle 6$ \\
 \cline{2-16}
 & Avg. rank
 & $3.56$               & 
 & $3.56$               & 
 & $4.19$               & 
 & $\boldsymbol{2.75}$  & 
 & $3.56$               & 
 & $5.88$               & 
 & $4.50$               & \\
\cline{1-16}
\noalign{\vskip\doublerulesep}
\cline{1-16}
\parbox[c]{3mm}{\multirow{9}{*}{\rotatebox[origin=c]{90}{Example-based F1}}}
 & \textsc{birds}
 & $70.38$              & $\scriptstyle 4$
 & $\boldsymbol{72.42}$ & $\scriptstyle 1$
 & $68.00$              & $\scriptstyle 7$
 & $71.18$              & $\scriptstyle 2$
 & $69.68$              & $\scriptstyle 6$
 & $70.06$              & $\scriptstyle 5$
 & $70.96$              & $\scriptstyle 3$ \\
 & \textsc{emotions}
 & $58.19$              & $\scriptstyle 7$
 & $59.06$              & $\scriptstyle 6$
 & $64.56$              & $\scriptstyle 3$
 & $65.75$              & $\scriptstyle 2$
 & $61.90$              & $\scriptstyle 5$
 & $\boldsymbol{69.24}$ & $\scriptstyle 1$
 & $62.59$              & $\scriptstyle 4$ \\
 & \textsc{enron}
 & $52.86$              & $\scriptstyle 6$
 & $53.87$              & $\scriptstyle 4$
 & $53.73$              & $\scriptstyle 5$
 & $53.91$              & $\scriptstyle 3$
 & $54.76$              & $\scriptstyle 2$
 & $46.61$              & $\scriptstyle 7$
 & $\boldsymbol{56.46}$ & $\scriptstyle 1$ \\
 & \textsc{llog}
 & $22.51$              & $\scriptstyle 6$
 & $23.21$              & $\scriptstyle 5$
 & $23.28$              & $\scriptstyle 4$
 & $26.30$              & $\scriptstyle 2$
 & $19.36$              & $\scriptstyle 7$
 & $\boldsymbol{33.17}$ & $\scriptstyle 1$
 & $24.16$              & $\scriptstyle 3$ \\
 & \textsc{medical}
 & $81.68$              & $\scriptstyle 3$
 & $80.07$              & $\scriptstyle 4$
 & $85.51$              & $\scriptstyle 2$
 & $\boldsymbol{85.97}$ & $\scriptstyle 1$
 & $53.14$              & $\scriptstyle 7$
 & $71.34$              & $\scriptstyle 5$
 & $64.72$              & $\scriptstyle 6$ \\
 & \textsc{scene}
 & $65.90$              & $\scriptstyle 7$
 & $71.69$              & $\scriptstyle 5$
 & $80.30$              & $\scriptstyle 2$
 & $\boldsymbol{81.69}$ & $\scriptstyle 1$
 & $70.30$              & $\scriptstyle 6$
 & $79.72$              & $\scriptstyle 3$
 & $74.23$              & $\scriptstyle 4$ \\
 & \textsc{slashdot}
 & $40.94$              & $\scriptstyle 5$
 & $44.04$              & $\scriptstyle 4$
 & $54.78$              & $\scriptstyle 2$
 & $\boldsymbol{58.62}$ & $\scriptstyle 1$
 & $32.57$              & $\scriptstyle 7$
 & $54.01$              & $\scriptstyle 3$
 & $39.15$              & $\scriptstyle 6$ \\
 & \textsc{yeast}
 & $61.56$              & $\scriptstyle 6$
 & $61.03$              & $\scriptstyle 7$
 & $62.78$              & $\scriptstyle 3$
 & $\boldsymbol{63.41}$ & $\scriptstyle 1$
 & $62.54$              & $\scriptstyle 4$
 & $61.71$              & $\scriptstyle 5$
 & $62.92$              & $\scriptstyle 2$ \\
 \cline{2-16}
 & Avg. rank
 & $5.50$               & 
 & $4.50$               & 
 & $3.50$               & 
 & $\boldsymbol{1.63}$  & 
 & $5.50$               & 
 & $3.75$               & 
 & $3.63$               & \\
\hline
\end{tabular}
\end{table}

We also conducted experiments on eight benchmark data sets from the Mulan and MEKA projects.\footnote{Data sets are available at \url{http://mulan.sourceforge.net/datasets-mlc.html} and \url{https://sourceforge.net/projects/meka/files/Datasets}.} As baselines we considered binary relevance (BR), label powerset (LP) and classifier chains (CC) \cite{read2009}, using XGBoost \cite{chen2016} as the base classifier. For BR and CC we used the logistic loss, for LP we used the softmax objective. While we tuned the number of rules $\numclassifiers \in \left \{ 50, 100, \dots, 10000 \right \}$ for our algorithm, XGBoost comes with an integrated method for determining the number of trees. The learning rate and the $L_2$ regularization weight were tuned in the same value ranges for both. Moreover, we configured XGBoost to sample 66\% of the training examples at each iteration and to choose from $\lfloor \log_2 \left( \numattr - 1 \right) + 1 \rfloor$ random attributes at each split. As classifier chains are sensitive to the order of labels, we chose the best order among ten random permutations. According to their respective objectives, the BR baseline is tuned with respect to Hamming, LP and CC are tuned with respect to subset 0/1 loss.

In \tab~\ref{tab:evaluation_benchmark}, we report the predictive performance of our methods and their competitors in terms of Hamming and subset 0/1 loss. For completeness, we also report the example-based F1 score (see, e.g., \cite{tsoumakas2010}). The Friedman test indicates significant differences for all but the Hamming loss. The Nemenyi post-hoc test yields critical distances between the average ranks of $2.91 / 3.19$ for $\alpha = 0.1 / 0.05$.

On average, \textsl{ex.w.-log.~multi} ranks best in terms of subset 0/1 loss. It is followed by LP, its counterpart \textsl{ex.w.-log.~single} and CC. As all of them aim at minimizing subset 0/1 loss, it is expected that they rank better than their competitors which aim at the Hamming loss. Most notably, since example-wise optimized multi-label rules achieve better results than single-label rules on all data sets (statistically significant with $\alpha = 0.1$), we conclude that the ability to induce such rules, which is a novelty of the proposed method, is crucial for minimizing subset 0/1 loss.

On the other hand, in terms of Hamming loss, rules that minimize the label-wise logistic loss are competitive to the BR baseline, without a clear preference for single- or multi-label rules. Interestingly, although the example-wise logistic loss aims at minimizing subset 0/1 loss, when using multi-label rules, it also achieves remarkable results w.r.t.\ Hamming loss on some data sets and consequently even ranks best on average.

\section{Conclusion}
\label{sec:conclusions}

In this work, we presented an instantiation of the gradient boosting framework that supports the minimization of non-decomposable loss functions in multi-label classification. Building on this framework, we proposed an algorithm for learning ensembles of single- or multi-label classification rules. Our experiments confirm that it can successfully target different losses and is able to outperform conventional state-of-the-art boosting methods on data sets of moderate size.

While the use of multivariate loss functions in boosting has not received much attention so far, our framework could serve as a basis for developing algorithms specifically tailored to non-decomposable loss functions, such as F1 or Jaccard, and their surrogates. The main drawback is that the computation of predictions for a large number of labels $n$ is computationally demanding in the non-decomposable case\,---\, solving the linear system has complexity $\mathcal{O} \left( n^3 \right)$. To compensate for this, we plan to investigate approximations that exploit the sparsity in label space. As the training complexity in MLC not only increases with the number of examples and attributes, but also with the number of labels, such optimizations are generally required for handling very large data sets.

\subsubsection*{Acknowledgments}

This work was supported by the German Research Foundation (DFG) under grant number 400845550. Computations were conducted on the Lichtenberg high performance computer of the TU Darmstadt.

\bibliography{bibliography}
\bibliographystyle{splncs04}

\end{document}